%% file: main.tex
\documentclass[letterpaper,10 pt,conference]{IEEEtran}
\IEEEoverridecommandlockouts
\IEEEsettopmargin{t}{72pt}

\usepackage{amsmath,amssymb,amsfonts}
\usepackage{graphicx}
\usepackage{textcomp}
\usepackage{xcolor}
\usepackage{lipsum}
\usepackage{svg}
\usepackage{comment}

\usepackage{subfigure}

\def\BibTeX{{\rm B\kern-.05em{\sc i\kern-.025em b}\kern-.08em
    T\kern-.1667em\lower.7ex\hbox{E}\kern-.125emX}}

\include{include}
\include{acronym}

\newcommand\copyrighttext{%
  \footnotesize \textcopyright 2024 IEEE. Personal use of this material is permitted.  Permission from IEEE must be obtained for all other uses, in any current or future media, including reprinting/republishing this material for advertising or promotional purposes, creating new collective works, for resale or redistribution to servers or lists, or reuse of any copyrighted component of this work in other works.
  }
\newcommand{\copyrightnotice}{%
\begin{tikzpicture}[remember picture,overlay]
\node[anchor=south,yshift=10pt] at (current page.south) {\fbox{\parbox{\dimexpr\textwidth-\fboxsep-\fboxrule\relax}{\copyrighttext}}};
\end{tikzpicture}%
}

\definecolor{somegray}{rgb}{0.5, 0.5, 0.5}
\newcommand{\darkgrayed}[1]{\textcolor{somegray}{#1}}
\makeatletter
\newcommand*\titleheader[1]{\gdef\@titleheader{#1}}
\AtBeginDocument{%
  \let\st@red@title\@title
  \def\@title{%
    \vskip-2.0em
    \bgroup\normalfont\large\centering\@titleheader\par\egroup
    \vskip0.0em\st@red@title}
}

\makeatother

\titleheader{\darkgrayed{This paper has been accepted for publication in the EMBC 2024 conference\\\copyright 2024 IEEE.}}

\title{
BISeizuRe: BERT-Inspired Seizure Data Representation to Improve Epilepsy Monitoring
}

 \author{\IEEEauthorblockN{
    Luca Benfenati\IEEEauthorrefmark{3},
    Thorir Mar Ingolfsson\IEEEauthorrefmark{1},
    Andrea Cossettini\IEEEauthorrefmark{1},\\
    Daniele Jahier Pagliari\IEEEauthorrefmark{3},
    Alessio Burrello\IEEEauthorrefmark{2}\IEEEauthorrefmark{3},
    Luca Benini\IEEEauthorrefmark{1}\IEEEauthorrefmark{2}}
     
    \IEEEauthorblockA{\IEEEauthorrefmark{1}Integrated Systems Laboratory, ETH Z{\"u}rich, Z{\"u}rich, Switzerland, \IEEEauthorrefmark{2}DEI, University of Bologna, Bologna, Italy}\IEEEauthorblockA{\IEEEauthorrefmark{3}DAUIN, Politecnico di Torino, Italy}

    \thanks{This project was supported by the Swiss National Science Foundation (Project PEDESITE) under grant agreement 193813.}
    
    }

\begin{document}
\thispagestyle{empty}
\pagestyle{empty}

\maketitle
\copyrightnotice

\vspace{-0.9em}
\begin{abstract}
This study presents a novel approach for EEG-based seizure detection leveraging a BERT-based model. The model, BENDR, undergoes a two-phase training process. Initially, it is pre-trained on the extensive Temple University Hospital EEG Corpus (TUEG), a 1.5 TB dataset comprising over 10,000 subjects, to extract common EEG data patterns. 
Subsequently, the model is fine-tuned on the CHB-MIT Scalp EEG Database, consisting of 664 EEG recordings from 24 pediatric patients, of which 198 contain seizure events. Key contributions include optimizing fine-tuning on the CHB-MIT dataset, where the impact of model architecture, pre-processing, and post-processing techniques are thoroughly examined to enhance sensitivity and reduce false positives per hour (FP/h). We also explored custom training strategies to ascertain the most effective setup. The model undergoes a novel \textit{second pre-training} phase before subject-specific fine-tuning, enhancing its generalization capabilities.
The optimized model demonstrates substantial performance enhancements, achieving as low as 0.23 FP/h, 2.5$\times$ lower than the baseline model, with a lower but still acceptable sensitivity rate, showcasing the effectiveness of applying a BERT-based approach on EEG-based seizure detection.

\indent \textit{Clinical relevance}— The model enhances clinical seizure detection, offering personalized treatments and better generalization to new patients, akin to successes with transformer-based models, thus significantly improving patient safety and care. 
\end{abstract}

\begin{IEEEkeywords}
Healthcare, EEG, Time Series Classification, Deep learning
\end{IEEEkeywords}

\input{sections/01_introduction}
\input{sections/02_methods}

\input{sections/03_results}

\section{Conclusion}\label{ch:conclusion}
This work proposes a BERT-based approach for seizure detection on the CHB-MIT dataset, demonstrating the potential of transfer learning from general \gls{eeg} data to seizure detection task. We also validate the effectiveness of a Transformer-based architecture which, once pre-trained on a large amount of unlabelled data, can partially overcome data labelling bottleneck and improve the state-of-the-art results. We then extensively explore hyper-parameters and pre-/post-processing techniques to improve the model performance.
The best model found obtains 0.23 FP/h, detecting 72.58\% of seizures. Our future work will include compressing the model and reducing the EEG channels to explore the possibility of its deployment on a wearable device, as well as implementing artefacts detection and removal algorithms \cite{artifact_detection} to further improve performances towards the clinical implementation of such methods.

\bibliographystyle{IEEEtran}
\bibliography{bibliography}

\end{document}

%% file: include.tex
\usepackage{comment}

\usepackage{nohyperref}

\definecolor{matplotlib0}{HTML}{1f77b4}
\definecolor{matplotlib1}{HTML}{d62728}
\definecolor{matplotlib2}{HTML}{2ca02c}
\definecolor{matplotlib3}{HTML}{ff7f0e}
\definecolor{matplotlib4}{HTML}{9467bd}
\definecolor{matplotlib5}{HTML}{8c564b}
\definecolor{matplotlib6}{HTML}{e377c2}
\definecolor{matplotlib7}{HTML}{7f7f7f}
\definecolor{matplotlib8}{HTML}{bcbd22}
\definecolor{matplotlib9}{HTML}{17becf}

\usepackage{mathtools} 

\usepackage{booktabs}
\usepackage{multirow}
\usepackage{colortbl}
\usepackage{tablefootnote}
\usepackage{threeparttable}

\usepackage[acronym, style=super, nonumberlist]{glossaries}

\usepackage{pgfplots}
\definecolor{color0}{rgb}{0.12156862745098,0.466666666666667,0.705882352941177} 
\definecolor{color1}{rgb}{1,0.498039215686275,0.0549019607843137}
\definecolor{color2}{rgb}{0.172549019607843,0.627450980392157,0.172549019607843} 
\definecolor{color3}{rgb}{0.83921568627451,0.152941176470588,0.156862745098039} 
\definecolor{color4}{rgb}{0.580392156862745,0.403921568627451,0.741176470588235}
\definecolor{colorblue}{rgb}{0.12156862745098,0.466666666666667,0.705882352941177} 
\definecolor{colorgreen}{rgb}{0.172549019607843,0.627450980392157,0.172549019607843} 
\definecolor{colorred}{rgb}{0.83921568627451,0.152941176470588,0.156862745098039} 
\definecolor{colorblack}{rgb}{0,0,0} 
\definecolor{colororange}{rgb}{1,0.56,0} 
\usepgfplotslibrary{fillbetween}
\usepgfplotslibrary{colormaps}
\pgfplotsset{compat=1.16}

\pgfplotscreateplotcyclelist{matplotlib}{
  {matplotlib0},
  {matplotlib1},
  {matplotlib2},
  {matplotlib3},
  {matplotlib4},
  {matplotlib5},
  {matplotlib6},
  {matplotlib7},
  {matplotlib8},
  {matplotlib9}
}

\pgfplotsset{every axis/.append style={
    cycle list name=matplotlib,
}}

\usepackage{listings}

\definecolor{code_default}{HTML}{000000}
\definecolor{code_keyword}{HTML}{AC4142}
\definecolor{code_identifier}{HTML}{D28445}

\lstdefinelanguage{RISCV}{
  sensitive=false,
  morecomment=[l]{//},
  alsoletter={.},
  morekeywords=[1]{
    lp.setup, mv, lw, p.lw, sw, p.sw, pv.sdotsp.b, pv.shuffle2.b, p.subNR, p.addNR
  },
  morekeywords=[2]{
    zero, ra, sp, gp, tp, t0, t1, t2, t3, t4, t5, t6, s0, s1, a0, a1, a2, a3, a4, a5, a6, a7, a8, a9, a10, a11,
  },
  morestring=[b]",
  morestring=[b]',
}[strings, comments, keywords]

\lstdefinestyle{RISCV_STYLE}{
  language=RISCV,
  numbers=none,
  basicstyle=\scriptsize\ttfamily\color{code_default},
  keywordstyle=[1]\color{matplotlib0},
  keywordstyle=[2]\color{matplotlib1},
  float,
  captionpos=b,
  belowskip=-0.5cm
}

\usepackage{algorithm}
\usepackage{algpseudocode}
\usepackage{float}
\newfloat{algorithm}{t}{top}


%% file: acronym.tex
\newacronym{simd}{SIMD}{Single Instruction, Multiple Data}
\newacronym{elu}{ELU}{Exponential Linear Unit}
\newacronym{relu}{ReLU}{Rectified Linear Unit}
\newacronym{rpr}{RPR}{Random Partition Relaxation}
\newacronym{mac}{MAC}{Multiply Accumulate}
\newacronym{dma}{DMA}{Direct Memory Access}
\newacronym{bmi}{BMI}{Brain--Machine Interface}
\newacronym{bci}{BCI}{Brain--Computer Interface}
\newacronym{smr}{SMR}{Sensory Motor Rythms}
\newacronym{eeg}{EEG}{Electroencephalography}
\newacronym{svm}{SVM}{Support Vector Machine}
\newacronym{svd}{SVD}{Singular Value Decomposition}
\newacronym{evd}{EVD}{Eigendecomposition}
\newacronym{iir}{IIR}{Infinite Impulse Response}
\newacronym{fir}{FIR}{Finite Impulse Response}
\newacronym{fc}{FC}{Fabric Controller}
\newacronym{nn}{NN}{Neural Network}
\newacronym{mrc}{MRC}{Multiscale Riemannian Classifier}
\newacronym{flop}{FLOP}{Floating Point Operation}
\newacronym{sos}{SOS}{Second-Order Section}
\newacronym{ipc}{IPC}{Instructions per Cycle}
\newacronym{tcdm}{TCDM}{Tightly Coupled Data Memory}
\newacronym{fpu}{FPU}{Floating Point Unit}
\newacronym{fma}{FMA}{Fused Multiply Add}
\newacronym{alu}{ALU}{Arithmetic Logic Unit}
\newacronym{dsp}{DSP}{Digital Signal Processing}
\newacronym{gpu}{GPU}{Graphics Processing Unit}
\newacronym{soc}{SoC}{System-on-Chip}
\newacronym{mi}{MI}{Motor-Imagery}
\newacronym{csp}{CSP}{Commmon Spatial Patterns}
\newacronym{fbcsp}{FBCSP}{Filter-Bank \acrlong{csp}}
\newacronym{pulp}{PULP}{parallel ultra-low power}
\newacronym{soa}{SoA}{state-of-the-art}
\newacronym{bn}{BN}{Batch Normalization}
\newacronym{isa}{ISA}{Instruction Set Architecture}
\newacronym{ecg}{ECG}{Electrocardiogram}
\newacronym{mcu}{MCU}{microcontroller}
\newacronym{rnn}{RNN}{recurrent neural network}
\newacronym{cnn}{CNN}{convolutional neural network}
\newacronym{tcn}{TCN}{temporal convolutional network}
\newacronym{emu}{EMU}{epilepsy monitoring unit}
\newacronym{dnn}{DNN}{deep neural network}
 \newacronym{bendr}{BENDR}{BERT-inspired Neural Data Representations}
\newacronym{bert}{BERT}{Bidirectional Encoder Representations from Transformers}
\newacronym{loocv}{LOOCV}{Leave-One-Out Cross-Validation}
\newacronym{smote}{SMOTE}{Synthetic Minority Oversampling TEchnique}
\newacronym{sswce}{SSWCE}{Sensitivity-Specificity Weighted Cross-Entropy}


%% file: sections/01_introduction.tex
\section{Introduction \& Related Works}\label{ch:introduction}
Epilepsy is a neurological disorder that affects over 50 million individuals in the world \cite{who_epilepsy} and is characterized by abnormal electrical activity of the brain that causes recurrent seizures. While medication management is the cornerstone of treatment, drug-resistant cases often require more advanced interventions, including surgery or neurostimulation. In this context, noninvasive \gls{eeg} data plays a crucial role in seizure detection and monitoring to trigger closed-loop actions such as neurostimulation.

Current methods of classifying raw \gls{eeg} mostly rely on \glspl{dnn}, which demonstrated higher accuracy than classical machine learning algorithms \cite{dnn_over_ml}. On the other hand, \glspl{dnn}s often face challenges when extracting features from relatively long time windows because of the lack of exploring the global correlation of all the input samples \cite{shallow1, shallow4}. Recent approaches \cite{eegformer, towards, epidenet} showed promising results when working on a reduced number of EEG channels for the deployment of seizure detection approaches on wearable low-power devices. However, most of the works mentioned above fail to reach a False Positive per hour (FP/h) ratio that can enable the clinical implementation of such methods. Although they can correctly detect almost all seizure events, they show a FP/h ratio that jeopardizes their real-world deployment, where wrongly reported seizures will alert the patient and reduce the detection device reliability. 
Furthermore, another key open challenge in all DNN models is the unavailability of very large high-quality labelled datasets \cite{scarcity_labelled_data}, as collecting and labelling seizure data is expensive and requires complex, human labour-intensive protocols, as well as patient hospitalization.

Our work investigates whether a model based on the Transformer architecture, which has recently gained widespread adoption in numerous deep learning applications, coupled with self-supervision and pre-training on large unlabeled datasets, can overcome the data labelling bottleneck and improve the state-of-the-art EEG-based seizure detection for long-term epilepsy monitoring. Our work focuses on the reduction of the FP/h ratio, even at the expense of a reduced sensitivity, working towards the objective of clinical deployment of seizure detection methods: we want a model that predicts a sufficient number of seizure events, with the least number of false alarms, thus reducing stress and anxiety of the monitored patients. The state-of-the-art in this domain is dominated by CNN-based methods \cite{cnnbased_1, sota1, cnnbased_3, cnnbased_4}, with the best approach achieving $100\%$ sensitivity and $0.58$ FP/h on CHB-MIT dataset \cite{chbmit}.
We take inspiration from a recent study~\cite{bendr} that applied Transformers and self-supervised sequence learning in \gls{bci} with \gls{eeg} data.
Our contributions are:
\begin{itemize}
    \item The adaptation for a seizure detection task on CHB-MIT of \gls{bendr} \cite{bendr}, an approach inspired by wav2vec 2.0 \cite{wav2vec} and \gls{bert} \cite{bert} which allows to exploit massive amounts of unlabeled \gls{eeg} data.
    \item An extensive task-specific optimization of the base BENDR model, including i) custom training strategies to improve seizure-detection fine-tuning on CHB-MIT; ii) the evaluation of the impact of architectural choices; iii) the application of pre- and post-processing techniques.
    \item A promising application of a transformer-based approach, pre-trained on a large amount of unlabelled data, for the seizure detection task.
\end{itemize}
Thanks to our optimizations, our best model can reduce the FP/h to as low as 0.23, outperforming the best SoA model by 2.5$\times$, with a lower yet acceptable sensitivity of 72.58\% \cite{acceptable_sensitivity}.

%% file: sections/02_methods.tex
\section{Materials \& Methods}\label{ch:methods}

\subsection{Model training and base architectures}

\subsubsection{Self-supervised pre-training}
The first model training step considered in this work consists in the same self-supervised pre-training scheme of \gls{bendr}, which in turn closely mirrors wav2vec 2.0~\cite{wav2vec}. Namely, the model is pre-trained on a large unlabelled \gls{eeg} dataset (TUEG)~\cite{tueg} to learn intrinsic patterns in EEG data. This initial phase then lays the groundwork for subsequent fine-tuning on our smaller and task-specific labelled dataset for seizure detection (CHB-MIT).  
Fig.~\ref{fig:bendr_architecture}a illustrates the model architecture used in the self-supervised pre-training phase.
The encoder's objective is to reconstruct the original sequence elements. This is achieved thanks to a contrastive loss function, which aims to align each element of the transformer's output sequence $c_i$ with the corresponding input $b_i$, notwithstanding the masking.
In our work, we do not re-implement this phase, and directly leverage the pre-trained weights of the original \gls{bendr} paper~\cite{bendr}.

\subsubsection{Fine-tuning}\label{subsec:fine-tune}
\begin{figure}
    \centering
    \includegraphics[width=\columnwidth]{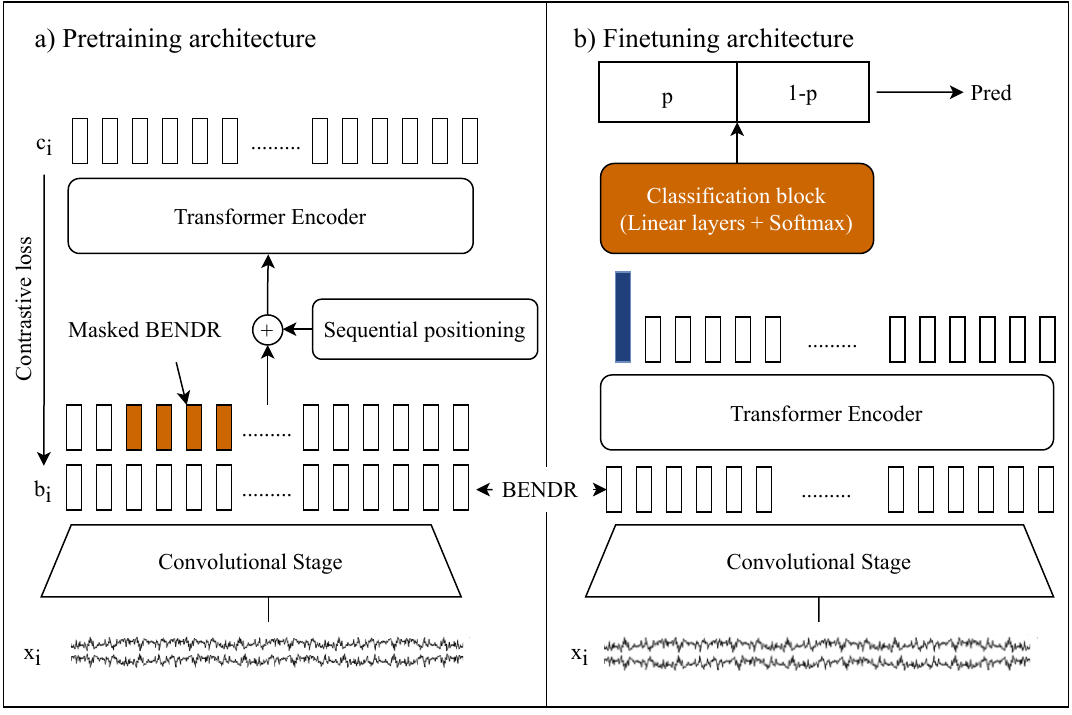}
    \caption{Model architecture used during (a) self-supervised pretraining (b) the second supervised pre-training and subject-specific fine-tuning.}
    \label{fig:bendr_architecture}
    \vspace{-0.25cm}
\end{figure}

The masking component is omitted in the fine-tuning phase, and \gls{bendr} vectors are fed directly to the Transformer. A classification block composed of one or more linear layers with a final Softmax activation is then employed to predict seizure occurrences based on the first element of the transformer output sequence.  Notably, during pre-training, this first input token was assigned a special value of $-5$ (selected to be out-of-distribution), and excluded from the contrastive loss. This step is crucial to ease downstream specialization to identify seizures accurately.
Fig.~\ref{fig:bendr_architecture}b shows the architecture used in this phase.

A significant modification of the fine-tuning task with respect to \gls{bendr} is the use of the \gls{sswce} loss, introduced by \cite{epidenet}.

Building upon the insights from prior research \cite{loocv_2}, this study employs a subject-specific approach for fine-tuning the seizure detection model. 
Such a tailored approach is pivotal in capturing the unique \gls{eeg} patterns and seizure characteristics inherent to each individual, enhancing sensitivity and specificity.
Precisely, we employ a \gls{loocv} strategy, which involves training the model on all seizure-containing records of a patient except one. The excluded record is then used as test set,  to assess the model's efficacy. A further 20\% of the training records is extracted to obtain a validation set, used for early stopping, learning rate scheduling, etc. The whole procedure is repeated cyclically, considering different records as test set.

\subsubsection{Second Supervised Pre-training}\label{subsec:second-pretrain}
\begin{figure}[h]
    \centering
    \includegraphics[width=0.9\columnwidth]{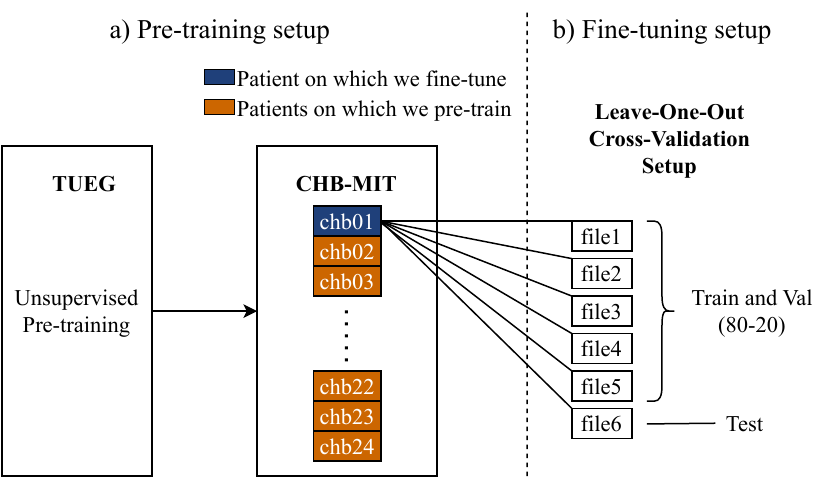}
    \caption{The model training scheme considering the pre-training, the second pre-training and the subject-specific \gls{loocv} fine-tuning step}
    \label{fig:second_pretraining} 
    \vspace{-0.25cm}
\end{figure}

A novel aspect of this work is the implementation of a \textit{second pre-training} phase on CHB-MIT, prior to the subject-specific \gls{loocv}. In this phase, the model is trained in a supervised way (as described in Sec.~\ref{subsec:fine-tune} and with the architecture of Fig.~\ref{fig:bendr_architecture}b), to predict seizures on \textit{all subjects except the target} one. This step aims to imbue the model with a broader understanding of \gls{eeg} patterns associated with seizures across different subjects, before honing in on the specific characteristics of a single patient. Figure \ref{fig:second_pretraining} illustrates the comprehensive training approach employed in the study.

\subsection{Task-specific optimizations}\label{sec:fine-tuning}
In order to maximize seizure detection accuracy, we explore with varying the model architecture, and apply several pre- and post-processing optimizations.

Keeping the fundamental architecture of Fig.~\ref{fig:bendr_architecture}, we explored different \textbf{classification block architectures}, finally converging to a sequence of 4 fully connected layers with decreasing dimensions, progressively reducing the feature space \{(512, 256), (256, 128), (128, 64), (64, 2)\} and improving the discriminative power of the model. We then \textbf{vary the number of blocks} in the Convolutional stage in \{3, 6\}, and the number of layers and heads in the Transformer Encoder in \{2, 4, 6, 8, 12\} to maximize seizure detection accuracy.
Additionally, we evaluate different \textbf{weights initialization policies} \cite{initialization_policy}, in order to apply weights pre-trained on TUEG to models of different sizes (e.g., with a smaller/larger number of layers).
Namely,  to initialize the additional layers of a bigger model with respect to the original one, we tested both the duplication of the pre-trained weights and random initialization. On the other hand, when testing a smaller model, we initialize its weights only with the layers shared with the original model.

Based on the hypothesis that early layers may capture the underlying, task-independent structure of EEG data, we consider \textbf{freezing the weights} of convolutional layers and fine-tuning only the Transformer encoder. For completeness, we also test how the model behaves when the Transformer encoder is frozen as well.

As \textbf{pre-processing}, we consider applying a  $5^{th}$-order Butterworth bandpass filter within a 0.5–50 Hz frequency range to the input, to smooth the signal and reduce ripple effects, followed by either MinMax or MeanStd normalization.
Moreover, to address the challenge posed by the highly unbalanced nature of EEG data, we tested two \textbf{oversampling} methods during supervised training phases, \gls{smote} and Weighted Random Sampler.

Lastly, we apply \textbf{post-processing} on the model's output to further reduce FP/h. Specifically, we go through the predicted labels with a sliding window and replace the central element with an aggregate prediction over the window. We consider two aggregation criteria: majority voting and minPooling. In the latter, we select the smallest predicted value in the window as the output. Windows of lengths 3, 5, 7 were considered.

\subsection{Training Protocol}\label{sec:training_details}
We train our models using an Adam optimizer with a learning rate of $1e^{-4}$ and $0.01$ weight decay, reducing the learning rate of factor 0.1 when the validation loss stops improving for $5$ consecutive epochs. We apply early stopping on the validation loss with a patience of $15$ epochs. Dropout layers with 50\% probability are added both in the Convolutional stage (between 1D convolutions and Group Normalization layers) and in the Transformer encoder to reduce overfitting. 

\subsection{Datasets}
\subsubsection{Self-supervised Pre-training Data}
For the self-supervised pre-training phase, the Temple University Hospital \gls{eeg} Corpus (TUEG) \cite{tueg} was utilized, encompassing 1.5 TB of clinical \gls{eeg} recordings. This dataset includes over 10,000 individuals, with diverse demographics, including 51\% females and a wide age range. Pre-existing pre-training weights were employed as the starting point for model initialization.
\subsubsection{Supervised Fine-tuning Data}
The supervised fine-tuning phase utilized the CHB-MIT Scalp \gls{eeg} Database \cite{chbmit}, comprising \gls{eeg} recordings from 24 pediatric subjects with intractable seizures at Boston Children's Hospital. The dataset, collected using the International 10-20 system, includes 664 EDF files, with 198 containing seizure events. For this study, only files with seizure occurrences were considered. Additionally, only 20 out of 23 channels were considered to match the channels used in the pre-training task on TUEG. All signals were sampled at 256 samples per second with 16-bit resolution and a window length of 8s was considered, without overlap.

%% file: sections/03_results.tex
\section{Experimental Results}\label{ch:results}
All results that are presented were obtained considering the LOOCV setup described in \ref{subsec:second-pretrain}, oversampling training data with a Weighted Random sampler. In terms of metrics, we focus on sensitivity and FP/h: sensitivity represents the ratio of correctly identified seizure \textit{events}, i.e., an event for which at least one of the segments that compose it is assigned a positive label; FP/h are the number of false alarms in an hour, which is inversely proportional to the specificity. After an extensive hyperparameters search for the \gls{sswce} loss, we found $\alpha=0.8$ and $\beta=0.2$ to be the best trade-off to maximize sensitivity and reduce FP/h.

\subsection{Model Performance}
Table \ref{tab:all_results} details our seizure detection results on the CHB-MIT dataset. First, we report the baseline models, i.e., the ones obtained fine-tuning on CHB-MIT the original BENDR architecture \cite{bendr} without any of our optimizations. This serves as a starting reference to demonstrate the improvements achieved by more refined solutions. Baseline results are reported both with pre-trained weights from TUEG and without any pre-training.
Without pre-training, the baseline achieves a 50.15\% sensitivity and an extremely high FP/h of 132.24. This is reduced to 10.38 FP/h when using pre-trained weights, demonstrating the crucial role of self-supervised pre-training for this kind of model.
In subsequent table rows, we report our successive refinements of the model obtained with optimizations described in the previous section, applied incrementally to the baseline. Results in bold represent the configurations that are used after each optimization step as a starting point for the next one.

First, we explore freezing the weights of the two main parts of the architecture: as expected, freezing the encoder weights worsens the results, while freezing the initial convolutional blocks leads to good generalization capabilities. We impute this to the fact that initial convolutional layers extract generic enough (i.e., not-task-specific) features from EEG signals during the pre-training. This model reaches better sensitivity (54.93\%) and FP/h (12.44).
On top of this, by applying our second (supervised) pre-training and filtering pre-processing, we obtain a model with an improved sensitivity of 69.11\% and a reduced FP/h of 6.95.
The last steps are the exploration of the model architecture and the application of pre- and post-processing. Firstly, MeanStd normalization behaves better than the MinMax one. From what concerns model architecture, we notice that a more complex classifier, can learn more discriminative EEG patterns and make more accurate predictions, reaching 81.87\% sensitivity and 2.75 FP/h. Then, we tune the number of convolutional layers and transformer block, obtaining the best generalization capabilities with 6 convolutional blocks and 4 attention layers in the encoder, with 4 heads each.  Our best post-processing pipeline applies minPooling on the output, using a sliding window of length $3$. This leads to 72.58\% of seizures detected, with 0.23 FP/h.

\begin{table}[]
\centering
\caption{Results obtained on optimization dimensions on CHB-MIT}
\label{tab:all_results}
\begin{tabular}{ccccc} \hline
\textbf{\begin{tabular}[c]{@{}c@{}}Optimization\\ Dimension\end{tabular}} &
  \multicolumn{2}{c}{\textbf{Setup}} &
  \textbf{\begin{tabular}[c]{@{}c@{}}Detected\\ Seizures [\%]\end{tabular}} &
  \textbf{FP/h} \\ \hline
\multirow{2}{*}{Baseline} &
  \multicolumn{2}{c}{w/o pre-trained weights} &
  \multicolumn{1}{c}{50.15} &
  \multicolumn{1}{c}{132.24} \\
                                 & \multicolumn{2}{c}{w pre-trained weights}  &    \textbf{43.24}      & \textbf{10.38}           \\ \hline
\multirow{2}{*}{\begin{tabular}[c]{@{}c@{}}Weight \\ Initialization\end{tabular}} &
  \multicolumn{2}{c}{freeze conv. layers} &   \textbf{54.93} &  \textbf{12.44} \\ 
                                 & \multicolumn{2}{c}{freeze transf. encoder} &   57.96        & 91.56          \\ \hline
Pre-training                     & \multicolumn{2}{c}{w second-pretraining}   & \textbf{69.11} & \textbf{6.95} \\ \hline
\multirow{2}{*}{Pre-processing}  & \multicolumn{2}{c}{MeanStd + Filtering}    & \textbf{82.44}            & \textbf{4.01}           \\
                                 & \multicolumn{2}{c}{MinMax + Filtering}     & 79.12          & 11.32           \\ \hline
\multirow{5}{*}{\begin{tabular}[c]{@{}c@{}}Model \\ Architecture\end{tabular}} &
  \multicolumn{2}{c}{new classifier} & \textbf{81.87} &  \textbf{2.75} \\ \cline{2-5} 
 &
  \multirow{4}{*}{\begin{tabular}[c]{@{}c@{}}Dimension\\ (conv., transf.)\end{tabular}} &
  \multicolumn{1}{l}{6, 12} &  86.55 & 4.59   \\
                                 &                    & 6, 4                 &     \textbf{83.57} & \textbf{2.81}       \\
                                 &                    & 6, 2                 &    85.91       & 5.35           \\
                                 &                    & 3, 4                 & 82.01           & 3.88           \\ \hline
\multirow{3}{*}{Post-processing} & \multicolumn{2}{c}{Majority voting}        &  78.16    & 0.60\\
                                 & \multicolumn{2}{c}{minPooling}             &  \textbf{72.58}    & \textbf{0.23}       \\
                                 & \multicolumn{2}{c}{Majority + minPooling}  &  73.28    & 0.26    \\ \hline \vspace{-0.5cm}
\end{tabular}
\end{table}

\subsection{Comparison with State-of-the-Art}
Table \ref{tab:comparison_with_soa} shows a comparative analysis with the latest state-of-the-art works that address seizure detection on CHB-MIT considering all patients and using from 18 up to 23 EEG channels. We distinguish between CNN-based and Transformer-based approaches. Our approach outperforms all of them in terms of FP/h, at the cost of lower sensitivity, demonstrating the capability of modern transformer models to capture complex epileptic patterns in EEG. Our approach reduces FP/h of the best performing CNN-based model \cite{cnnbased_4} by 2.5$\times$, while simplifying the heavy pre-processing and feature extraction steps, typical of CNN-based approaches, at the expense of reduced sensitivity of the model (27.42\% decrease). Our approach significantly improves upon other transformer-based models, with a 4.7$\times$ lower FP/h with respect to \cite{transformerbased_2}. Noteworthy, this is crucial for a real-life closed-loop system, given that a high number of FP/h per hour would cause many warnings in patients, which in turn increase their stress, as discussed in~\cite{prioritize_fph}. Moreover, a sensitivity over $50\%$ has already been showed to be an acceptable requirement for a seizure detection method \cite{acceptable_sensitivity}.  

\begin{table}[]
\centering
\caption{Comparison with SoA seizure detection on CHB-MIT}
\label{tab:comparison_with_soa}
\begin{tabular}{cccc}
\hline
\textbf{Type}                      & \textbf{Work}                            & \textbf{\begin{tabular}[c]{@{}c@{}}Detected\\ Seizures [\%]\end{tabular}} & \textbf{FP/h}\\ \hline
\multirow{5}{*}{CNN-based}         & Qui et al. \cite{lightseizurenet}         & 97.1  & 7.51                         \\
 & Wang et al. \cite{cnnbased_1}    & 97.52 & 3.42                    \\
 & Zhao et al. \cite{sota1}        & 97.79 & 1.25               \\
 & Li et al. \cite{cnnbased_3}     & 98.47 & 0.63              \\
 & Sahani et al. \cite{cnnbased_4} & 100   & 0.58\\ \hline
\multirow{3}{*}{Transformer-based} & Yan et al. \cite{transformerbased_1}     & 96.01 & 27.14              \\
                                   & Hussein et al. \cite{transformerbased_2} & 99.8 & 1.08                   \\
 & Our work                        & \textbf{72.58} & \textbf{0.23} \\ \hline \vspace{-0.5cm}
\end{tabular}
\end{table}